# Data Augmentation Methods for Anaphoric Zero Pronouns

**Abdulrahman Aloraini**[1,2], **Massimo Poesio**[1]
[1]Queen Mary University of London, United Kingdom
[2]Qassim University, Saudi Arabia
{a.aloraini, m.poesio}@qmul.ac.uk

## Abstract

In pro-drop language like Arabic, Chinese, Italian, Japanese, Spanish, and many others, unrealized (null) arguments in certain syntactic positions can refer to a previously introduced entity, and are thus called anaphoric zero pronouns. The existing resources for studying anaphoric zero pronoun interpretation are however still limited. In this paper, we use five data augmentation methods to generate and detect anaphoric zero pronouns automatically. We use the augmented data as additional training materials for two anaphoric zero pronoun systems for Arabic. Our experimental results show that data augmentation improves the performance of the two systems, surpassing the state-of-the-art results.

## 1 Introduction

In pronoun-dropping (pro-drop) languages such as Arabic (Eid, 1983), Chinese (Li and Thompson, 1979), Italian (Di Eugenio, 1990) and other romance languages (e.g., Portuguese, Spanish), Japanese (Kameyama, 1985), and others (Kim, 2000), certain arguments can be omitted in which a pronoun is used in English: such arguments are called zero-pronouns (ZP) in NLP. Anaphoric zero pronouns (AZP) are a special case of zero pronouns. AZPs refer to one or more noun entities previously mentioned in a text, as in the following example from the Arabic portion of OntoNotes (in Arabic OntoNotes 5.0, ZPs are denoted as * in Arabic text).

.. المفارقة الأخرى عن بوش هي عدم حماسته للمؤتمر الدولي، اذ انه من البداية، يريد * اجتماعا مختلفا ....

*Ironically, Bush did not show any enthusiasm for the international conference, because since the beginning, (he) wanted to attend another conference ...*

In the example, the AZP, indicated with '*', refers to a singular masculine person that has been mentioned previously, Bush (بوش). English is not a pro-drop language (White, 1985) so the null argument in Arabic has to be expressed using an overt pronoun (he).

AZP resolution involves two tasks: identification and resolution. AZP identification is the task of identifying the locations where the pronoun was omitted (the * in the example). AZP resolution is the task of finding the antecedent that an AZP refers to (Bush in the example). Contextual information is required for these two tasks, such as that provided by the verb which has the AZP as an argument (in the example the verb is "wanted/يريد").

The AZP problem has inspired much research because AZP interpretation benefits many natural language processing (NLP) tasks such as machine translation (Mitkov and Schmidt, 1998). Although AZPs are common (Chen and Ng, 2016), they are not always annotated in NLP corpora. There are two reasons for this. First, AZPs have no surface realization and the focus in coreference is usually on arguments realized on the surface (Lee et al., 2005). Second, pro-drop languages have several types of ZPs, not all of which anaphoric, which can make it challenging to identify AZPs. Therefore, the number of datasets with annotated AZPs is small (Konno et al., 2020). We investigate therefore automatic methods for augmenting these existing data, applying them to Arabic. Our contributions are as follows:

- We apply various data augmentation methods to detect potential AZPs in unannotated sentences, and to generate sentences containing AZPs.

- We propose a method to automatically find the true antecedent of AZPs.

- The augmented data improve AZP identification and resolution for Arabic, and surpass the

current state-of-the-art results.

The rest of the paper is organized as follows. We discuss AZP and data augmentation related literature in Section 2. We explain our data-augmentation methods in Section 3. We discuss the evaluation settings and results in Section 4 and Section 5 respectively. We conclude in Section 6.

## 2 Related Work

### 2.1 Anaphoric Zero Pronouns

**Arabic**: There have been a few proposals devoted to AZP tasks and null arguments in general. Green et al. (2009) proposed a conditional-random-field (CRF) sequence classifier to detect Arabic noun phrases, and captured ZPs implicitly. Bakr et al. (2009) applied a statistical approach to detect empty categories. Gabbard (2010) proposed a pipeline made of maximum entropy classifiers which jointly make a CRF to retrieve Arabic empty categories. Aloraini and Poesio (2020b) proposed the first neural model for resolving Arabic AZP, but they did not consider the AZP identification step. Aloraini and Poesio (2020a) showed a multilingual approach to detect AZPs for Arabic and Chinese using BERT (Devlin et al., 2018).

**Other languages**: for Chinese, Converse (2006) studied AZP resolution and applied a rule-based approach that employed Hobbs algorithm (Hobbs, 1978) to resolve ZPs in the Chinese Treebank; however, did not attempt to automatically identify AZP. Yeh and Chen (2006) is another rule-based approach for Chinese which used a set of hand-engineered rules to identify and resolve AZPs. Zhao and Ng (2007) proposed the first machine learning approach to Chinese AZPs identification and resolution. They applied decision trees incorporated with a set of syntactic and positional features. Other proposals followed the machine learning approach targeting Chinese (Kong and Zhou, 2010; Chen and Ng, 2013, 2014, 2015, 2016; Yin et al., 2016, 2017; Liu et al., 2017; Chang et al., 2017; Yin et al., 2018; Kong et al., 2019). There has been also a great deal of research on identification and resolution of AZPs, particularly in Japanese (Yoshimoto, 1988; Kim and Ehara., 1995; Aone and Bennett, 1995; Seki et al., 2002; Isozaki and Hirao, 2003; Iida et al., 2006, 2007; Sasano et al., 2008, 2009; Sasano and Kurohashi, 2011; Yoshikawa et al., 2011; Hangyo et al., 2013; Iida et al., 2015; Yoshino et al., 2013; Yamashiro et al., 2018), but also in other languages, including Korean (Han, 2004; Byron et al., 2006), Spanish (Ferrández and Peral, 2000; Rello and Ilisei, 2009), Portuguese (Rello et al., 2012), Romanian (Mihăilă et al., 2011), Bulgarian (Grigorova, 2013), and Sanskrit (Gopal and Jha, 2017). Iida and Poesio (2011) proposed the first multilingual approach for AZP resolution.

These proposals applied their AZP systems to labeled dataset. Even though data augmentation improved many NLP tasks, as discussed in Section 2.2, none of the above proposals considered augmenting AZPs datasets automatically. As far as we know, (Konno et al., 2020) is the only proposal to use data augmentation for AZP resolution. Konno et al applied a data augmentation method called contextual data augmentation (CDA) to Japanese AZPs. CDA method is based on language models to generate different variants of a labeled input by masking certain tokens and then predict them. Our work is different in that we apply various methods to detect, generate, and resolve AZPs while they focus generating samples for AZP resolution. The generated AZP samples are of numerous types and they are automatically prepared for both AZP tasks.

### 2.2 Data Augmentation

Data augmentation is an active research topic and has been applied in different areas of research of NLP (Feng et al., 2021; Chen et al., 2021). However, there has been very limited proposals for AZP data augmentation. Zhang et al. (2015) augmented data by replacing a word with its synonyms to improve text classification. Sennrich et al. (2015) augmented data by translating a sequence from one language to another, and then translating the sequence back into the original language. The new data were used to enhance the performance of neural machine translation models. Wang et al. (2018) examined various methods for data augmentation and proposed to randomly replace words in source and target languages to improve neural machine translation. Şahin and Steedman (2019) removed dependency links and modified tree nodes to create an augmented dataset for Part-of-Speech tagging. Gulordava et al. (2018) replaced words with other words that share the same Part-of-Speech, morphological features, and dependency labels, to improve subject-verb agreement models. Feng et al. (2019) introduced a pipeline called SMERTI which combines various data augmentation methods, such as, entity replacement, similarity masking, and text in-filling. Grundkiewicz et al. (2019) used a spellchecker to augment training data which are then used to pre-train sequence-to-sequence models. Singh et al. (2019) introduced a cross-lingual data augmentation called XLDA, evaluated on 14 languages on a natural language inference (XNLI) benchmark and question-answering task. XLDA replaces segments of an inputs text with its translation in other languages. Kumar et al. (2019) proposed DiPS, a model that generates various paraphrased sentences used to train conversational agents and in text summarization tasks. Andreas (2019) proposed rule-based data augmentation, which replaces segments of inputs that share similar context to improve the training of n-gram and sequence-to-sequence language models. Guo et al. (2020) proposed a statistical approach called SeqMix to decide which token to use at each position of an input, and they also provided a framework that combines several data augmentation approaches for several NLP tasks. Chen et al. (2020) represented datasets as graphs and proposed methods to augment data based on graph theory for paraphrasing. Ding et al. (2020) trained a language model on the linearized version of the input to synthesize data for low-resource sequence-tagging. Feng et al. (2020) inserted character-level synthetic noise and word hypernyms to augment data for text generation. Louvan and Magnini (2020) used a set of augmentation methods that span words and modify sentences for slot filling and intent detection. (Ri et al., 2021) proposed a method to augment ZPs (not AZPs). They delete personal overt pronouns which results in extra sentences with no pronouns, and use them in Japanese-English translation. Other proposals investigated data augmentation for neural machine translation (Vaibhav et al., 2019; Gao et al., 2019; Nguyen et al., 2019), text classification (Kobayashi, 2018; Wei and Zou, 2019; Anaby-Tavor et al., 2020; Jindal et al., 2020), and question-answering (Kafle et al., 2017; Yang et al., 2019).

Data augmentation methods helped to improve various NLP tasks; however, very few proposals (Konno et al., 2020) considered the methods for AZP tasks which was focused on generating AZPs for resolution. This paper explores the direction of exploiting several data augmentation methods for AZP resolution and identification, and how they can benefit both tasks.

## 3 Methodology

We applied the following five methods to generate AZPs:

1. **OntoNotes Patterns (ONP):** AZPs may occur more frequently in the company of certain verbs. To find the most frequent collocations, we apply the t-test on the Part-of-Speech sequences of AZP sentences. We tried a window of one, two, and three and we found empirically a window of two to detect many correct AZP collocations.

2. **Removing Subject Mention (RSM):** AZPs are dropped subjects of verbs. When an (explicit) mention is the subject of a verb phrase, we remove the mention to obtain an AZP sentence.

3. **Masking Candidate Mentions (MCM),** also known as contextual data augmentation (CDA). We mask the true antecedents of AZPs and use a language model to find the semantically most similar words, which are then used to produce new sentences by replacing the original word.

4. **Back Translation (BT):** the AZP examples in OntoNotes are in Arabic. We translate them into English using the GoogleTrans API,[1] and then translate them back to Arabic.

5. **Changing Subject–verb Agreement (CSA):** the verb in an AZP construction and its true antecedent agree on number which can be in singular, dual, or plural form. We change their agreement number from one form to another. For example, if the AZP verb and its reference are in singular form, we change the verb and its reference to dual or plural form.

Table 1 shows the number of collected data of each method. AZPs interpretation involves two steps: AZP detection, and AZP resolution. The detection step finds if a sentence has AZPs while the resolution resolves AZPs to their true antecedent. The true antecedent might be present in the same AZP sentence or any previous sentences. In our experiments, we decided to collect AZP samples such that each sample has two sentences: the first has the true antecedent and the second has its AZP. While it is possible to automatically create AZP samples of one sentence only (the AZP position and its true antecedent in the same sentence), but two-sentence samples provide more context and information about AZPs.

---
[1] https://pypi.org/project/googletrans/

| Method | Training |
| --- | --- |
| OntoNotes Patterns (ONP) | 369 |
| Removing Subject (RSM) | 1,196 |
| Masking Candidate (MCM) | 736 |
| Back Translation (BT) | 501 |
| Changing Subject-Verb Agr. (CSA) | 104 |
| Total | 2,906 |

Table 1: The number of the augmented data of each method, and their total.

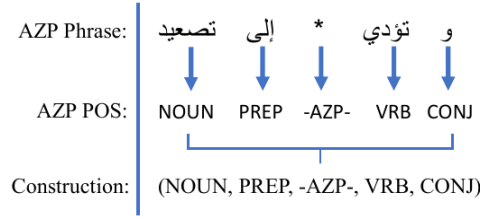

Figure 1: An example of one construction of an AZP from OntoNotes. For every AZP position, we extract the POS of two words before and two after, in order to create one construction.

### 3.1 OntoNotes Patterns

AZP sentences may occur more frequently in certain constructions. To find these constructions, we apply a window of one, two, and three tokens on the Part-of-Speeches of AZP samples. We found a window of two (two words before an AZP and two words after as shown in Figure 1) to provide many correct collocations. We count the frequencies of the constructions and apply the t-test score as in (Manning and Schutze, 1999) to see how probable a construction is:

$$t = \frac{\bar{x} - \mu}{\sqrt{\frac{s^2}{N}}}$$

Where $\bar{x}$ is the sample mean of a Part-of-Speech, and $s^2$ is its sample variance, N is its sample size, and $\mu$ is its distribution mean. We choose the highest five t-test scores and apply them to large public text datasets i.e Wikipedia. We tag the sentences in Wikipedia summary sections with their Part-of-Speech using CAMeL tools (Obeid et al., 2020) and apply the OntoNotes constructions on each sentence except the first. When one sentence matches a pattern, it suggests that the sentence might be an AZP. The AZP true antecedent is usually in the first sentence of the summary section. We join the first sentence and the detected sentence together to represent one AZP sample, as shown in Figure 3. In Section 3.6, we discuss in details why we follow this approach and the challenges of finding the true antecedent of an AZP.

### 3.2 Removing Subject Mentions

Subject nouns or pronouns of a verb phrase might be optional. Removing these subjects transforms a sentence into a null-subject, as shown in the following example:

تقع لندن على نهر التامز
*London is located on the Thames river.*

In the example, removing 'London' from the Arabic sentence changes it into an AZP sentence and grammatically correct. However, when the subject is removed, readers might not know what the AZP refers unless they have access to more context i.e previous sentences. Therefore, in our experiments each AZP sample has two sentences. The first sentence contains the mention that the AZP refers to, and the second contains the AZP gap.

### 3.3 Masking Candidate Mentions

We use the multilingual BERT[2] to mask antecedents of AZPs and replace them with their most similar semantic token. We replace the antecedent token with [MASK][3], and predict its most similar semantic word. We replace the original mention with the predicted mention from BERT. However, even though the predicted token might be semantically similar to the original token, it might not agree in number or gender with the AZP verb. For example, the antecedent in the original sample can be *student* and BERT replaces it with *students*. Such samples can distort the training for machine learning algorithms because it associates the morphemes with wrong morphological features. We address this problem in Section 3.7.

### 3.4 Back Translation

It has been shown that translating a sentence from one language to another, and then translating it back to the original language to be beneficial to some NLP tasks (Xie et al., 2018; Vaibhav et al., 2019;

---

[2] https://github.com/google-research/bert
[3] [MASK] is a special token in BERT which is used for prediction.

Zhang et al., 2019). Back Translation generates a paraphrased version of the original input adding noise, such as, semantic and syntactical changes. Therefore, we translate the Arabic samples to English, and translate them back to Arabic.

### 3.5 Changing Subject–verb Agreement

AZP verb and its antecedent usually agree on number, gender, and person. We change the verb and its antecedent number agreement from one form to another. For example, if the number of an AZP sample is singular, we change it to dual or plural. To find and generate verb inflections and mention number, we use CAMeL which consists of a set of NLP tools for Arabic.

The five methods we discussed are used create AZP samples, but they are applied differently. The first two (ONP and RSM) are used to detect AZPs on text while the other three (CSA, MCM, and BT) are used to generate AZP from existing AZP samples. Thus, we apply the two methods on the Wikipedia summary sections to initially collect AZPs, and then use the other three to generate extra samples. An illustration of all data augmentation methods and steps in Figure 2.

### 3.6 Finding the True Antecedent

An AZP refers to a preceding mention in a text. The mention can be in the same sentence as the AZP or in previous sentences. Finding the true mention in a sentence using patterns is not trivial because resolving AZPs involves reasoning, context, background knowledge of real world, and deep understanding of a language characteristic, such as, its morphology (Alnajadat, 2017). At the beginning of the experiment, we apply the two methods (ONP and RSM) on Arabic newspaper articles of Abu El-Khair corpus (El-Khair, 2017) to detect AZP locations. For resolving AZPs to their antecedents, we used AZP context features to find the true antecedent. The main features are the agreement morphemes between the AZP verb and its antecedent. The result was we managed to identify many AZP locations successfully; however, we encountered two problems when we tried finding their true antecedents. First, there were many mentions that share similar features as the true antecedent which led to many incorrect antecedent reference. Second, some locations of the true antecedents were very far from the AZP gap.

To alleviate the above-mentioned limitations of AZP resolution, we apply the two methods on Wikipedia articles. Every Wikipedia articles focuses on a single topic especially in the summary section. Summary sentences can contain AZPs and they usually refer to the Wikipedia title. Therefore, we apply the five methods on the summary sentences, and we consider the title to be the true antecedent of every AZP. The true antecedent always appears in the first sentence i.e the title of Wikipedia page or part of it. So for every AZP in the other sentences of the same page, they refer to the mention in the first sentence. So each AZP sample consists of the first sentence of the Wikipedia page and the sentence where it has the AZP. The reason why we apply the methods only on Wikipedia summary sections because their sentences usually focus on briefly describing the entity of the page. The sentences of other sections might have AZPs but they also include other mentions which can make it challenging to resolve the AZPs to their true antecedent. An illustration of creating one sample from Wikipedia is in Figure 3. After collecting the samples using ONP and RSM methods, we apply the other methods MCM, BT, CSA to generate extra samples.

### 3.7 Filtering Samples

After data creation and generation, we found some samples to be grammatically incorrect or not AZPs. For example, sometimes MCM method replaces the original antecedent of an AZP with a mention of a different number or gender. BT method can be problematic in some cases because it predicts the dropped subject which converts an AZP sample into a non-AZP sample. Another problem of BT method is that the AZP verb in the original sample is used to track the AZP location. If the AZP verb is translated back into another similar verb, the AZP location can not be retrieved back. To enhance the quality of the augmented data produced by the MCM method, we use Camel tools [4] which has a morphological analyser to find the gender and number of mentions. We use the analyser to see if the AZP verb and its antecedent agree on number and gender. For the grammatically incorrect cases, we remove samples that do not have complete agreements. For the added subject after the BackTranslation step, we remove the subjects in a similar way as we do in RSM method. If the verb of the original sample is absence after the BackTranslation, we remove the sample.

---

[4] https://github.com/CAMeL-Lab/camel_tools

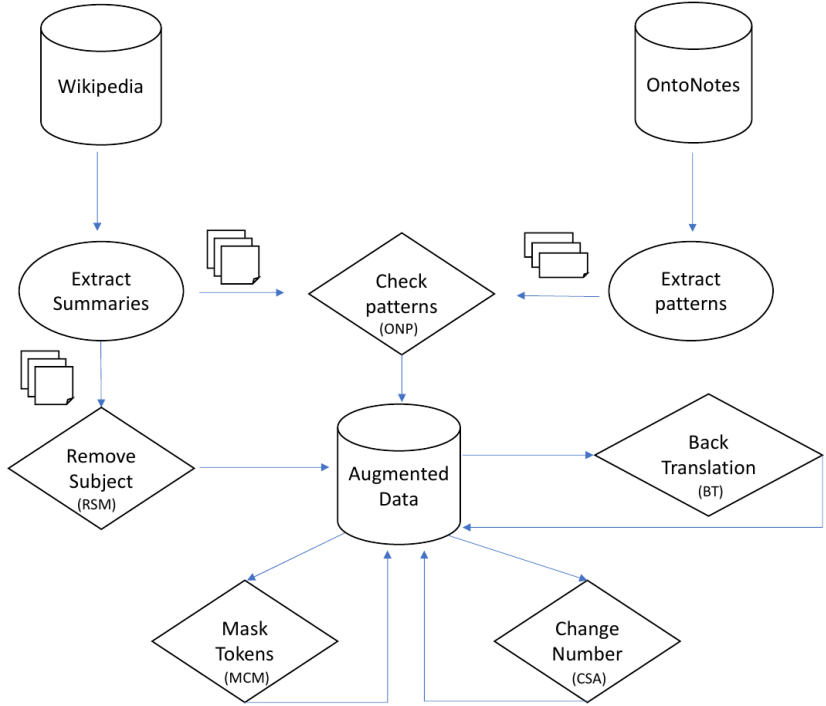

Figure 2: Overall picture of all data augmentation methods for Anaphoric Zero Pronouns. We extract patterns from OntoNotes and check them on Wikipedia summary sentences (ONP method). Removing Subject (RSM) is also applied to the summary sentences. after collecting AZP samples, We create extra data applying MCM, BT, and CSA methods.

## 4 Evaluation

### 4.1 Dataset

Most existing Arabic corpora do not have AZP annotated. As far as we know, OntoNotes 5.0 is the only source that has Arabic AZPs annotated. The distribution of documents, sentences, words, and AZPs in OntoNotes are in Table 2.

| Category | Training | Dev | Test |
| --- | --- | --- | --- |
| Documents | 359 | 44 | 44 |
| Sentences | 7,422 | 950 | 1,003 |
| Words | 264,589 | 30,942 | 30,935 |
| AZPs | 3,495 | 474 | 412 |

Table 2: Basic statistics about AZPs in Arabic OntoNotes. The total number of AZPs is 3495, 474, and 412 for training, development, and test respectively.

### 4.2 AZP systems

For AZP identification, we use the model by (Aloraini and Poesio, 2020a) which is a binary classifier that takes a candidate ZP location as input, and classifies whether it as an AZP or not. For AZP resolution, we use the model by (Aloraini and Poesio, 2020b) which combines BERT representations and additional task-related features of AZPS to learn their true antecedent. We adopt their systems and apply the same settings on Arabic AZPs. We then evaluate each method of our data augmentation for both systems to see its effect. We then try using all the collected data, and try various binary combinations as we will see in Section 5.

### 4.3 Data Preprocessing

Arabic is a morphologically rich language with a large set of morphological properties. Arabic text can suffer from sparsity (different forms for the same word) and ambiguity (same form for numerous words) if the text is not pre-processed. There are two reasons for these problems. First, certain letters can have different forms which are usually misspelled, such as the various forms of the letter "alif". Second, same word can have a fully diacritized, partially diacritized or undiacritized forms. Retaining diacritics can complex word representation and model training (Habash and Sadat, 2006). Pre-processing Arabic text improves the overall performance. For example, Aloraini et al. (2020) fol-lowed the pre-processing steps by Althobaiti et al. (2014) which led to a significant improvement in the performance of Arabic coreference resolution with an increase of 7% F1 score more than the baseline. Therefore, we follow their steps for pre-processing text which are:

- We normalize the various forms of the letter "alif" (ا,آ,أ,إ) to the letter "ا".

- We remove all diacritic marks on words, such that each word has its undiacritized form.

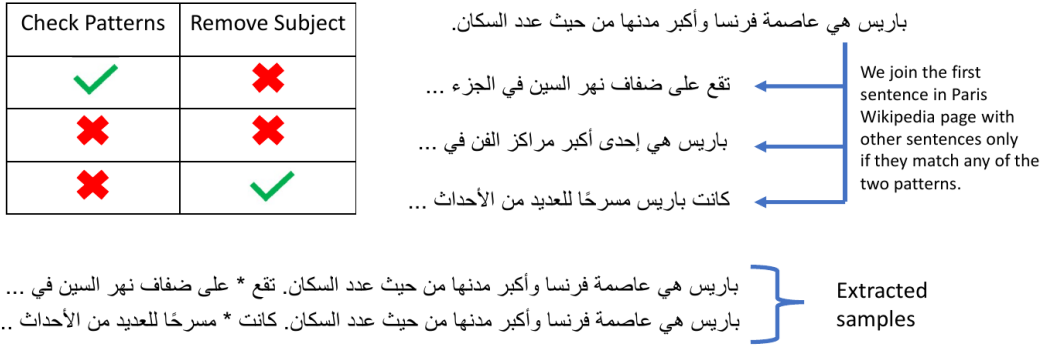

Figure 3: Illustration of finding AZPs using OntoNotes Patterns (ONP) and Remvoing Subject Mnetion (RSM) on Paris, from Arabic Wikipedia. Given a Wikipedia page, we join the first sentence with any other sentence in the summary only if they match any of the two data augmentation methods. The extracted samples are made of the first sentence and the matched sentences. The first sentence contains the true antecedent and the second the AZP.

| Original text | لَا بَأْسَ أَن تُقَالَ |
| --- | --- |
| pre-processed text | لا باس ان تقال |

Table 3: An example of how we pre-process Arabic text. We normalize all 'alif' letters forms, and remove all diacritics.

We show an example of an original and pre-processed sentence from OntoNotes in Table 3.

## 5 Results

We use the systems from (Aloraini and Poesio, 2020a) and (Aloraini and Poesio, 2020b) as baselines for AZP identification and resolution respectively. To see the effect of the collected samples, we use their systems following the same settings. We evaluate the augmented data of each method with different settings, and see their performance on the precision, recall, and F1 scores.

**AZP identification** in Table 4, we train the baseline system with every data augmentation method and evaluate on the test portion. We see the results vary. RSM and BT improve the scores with an increase of +0.4 and +0.1 compared to the baseline F1 score. ONP and CSA hurt the scores with a decrease of -0.3 and -0.7 compared to the baseline F1 score. When we train the system with ALL the augmented data, it outperforms the baseline with an increase of +0.3, but still less than when we use RSM by itself. RSM method provides the best performance and outperforms the baseline and all other settings scores.

**AZP resolution** we evaluate each augmented-data method and the results are shown in Table 5. Almost every data augmentation method increases the overall performance when it is trained with CoNLL-2012 except for CSA method. ONP, RSM, MCM, and BT increase the F1 score with +0.5, +1.6, +0.8, and +1.0 F1 scores respectively while CSA decreases the score by -0.2, compared with the baseline's F1 score. The best settings when we train ALL the augmented data with CoNLL-2012 which results an increase of +1.8.

**Combinations** we also try combining two methods together to see if they can outperform when we use every method separately. We show the AZP identification results in Table 6 and Table 7 for AZP resolution. As shown in the tables, some combinations can improve the scores; however, no single combination outperforms the highest scores in Tables 4 and 5.

### 5.1 Discussion

The augmented data helps to improve the results for both AZP identification and resolution. The improvements in scores for AZP resolution are more than AZP identification. To understand the reasons, we investigated the Arabic portion of CoNLL-2012 and the outputs of the two systems. In CoNLL-2012, we found that many AZP samples to be duplicates but they refer to different types of mentions. The augmented data has more variance of AZP verbs; however, many were not present in the test part which might have led the model to associate these newly seen AZP verbs to cases that are not AZPs in the test set. We also examined the identified and resolved AZPs, and we found several cases to be verbs or mentions that are not subset of the training or the augmented data. Even though the data augmentation methods we showed helped to alleviate this problem, we still need more annotated data and explore more data augmentation methods.

| Training settings | Test Evaluation | | | |
| --- | --- | --- | --- | --- |
| | P | R | F1 | diff |
| CoNLL-2012 [b] | **60.0** | 78.9 | 68.2 | - |
| + ONP | 59.4 | 79.3 | 67.9 | -0.3 |
| + RSM | 59.8 | **80.4** | **68.6** | **+0.4** |
| + MCM | 59.6 | 79.8 | 68.2 | 0 |
| + BT | 59.6 | 80.2 | 68.3 | +0.1 |
| + CSA | 59.3 | 78.5 | 67.5 | -0.7 |
| CoNLL-2012 + ALL | 59.8 | 80.2 | 68.5 | +0.3 |

Table 4: AZP identification training settings with each data augmentation technique, and their results on the test set. [b] is the baseline scores (Aloraini and Poesio, 2020a). *diff* is the difference between a method's F1 score and the baseline.

| Training settings | Test Evaluation | | | |
| --- | --- | --- | --- | --- |
| | P | R | F1 | diff |
| CoNLL-2012 [b] | 64.4 | 51.8 | 57.4 | - |
| + ONP | 64.8 | 52.4 | 57.9 | +0.5 |
| + RSM | 65.6 | 53.7 | 59.0 | +1.6 |
| + MCM | 65.3 | 52.6 | 58.2 | +0.8 |
| + BT | 65.3 | 52.9 | 58.4 | +1.0 |
| + CSA | 64.4 | 51.6 | 57.2 | -0.2 |
| CoNLL-2012 + ALL | **65.8** | **53.9** | **59.2** | **+1.8** |

Table 5: AZP resolution training settings with each data augmentation technique, and their results on the test set. [b] is the baseline scores (Aloraini and Poesio, 2020b). *diff* is the difference between a method's F1 score and the baseline.

| Training settings | Test Evaluation | | | |
| --- | --- | --- | --- | --- |
| | P | R | F1 | diff |
| CoNLL-2012 [b] | **60.0** | 78.9 | 68.2 | - |
| + ONP + RSM | 59.6 | 79.7 | 68.2 | 0.0 |
| + ONP + MCM | 59.5 | 79.7 | 68.1 | -0.1 |
| + ONP + BT | 59.8 | 79.8 | 68.3 | +0.1 |
| + ONP + CSA | 59.7 | 78.8 | 67.9 | -0.3 |
| + RSM + MCM | 59.8 | 79.0 | 68.1 | -0.1 |
| + RSM + BT | 59.8 | 79.8 | 68.4 | +0.2 |
| + RSM + CSA | 59.8 | **80.2** | **68.5** | **+0.3** |
| + MCM + BT | 59.8 | 79.7 | 68.3 | +0.1 |
| + MCM + CSA | 59.7 | 79.7 | 68.2 | 0.0 |
| + BT + CSA | 59.7 | 79.0 | 68.0 | -0.2 |

Table 6: AZP identification training settings with different combinations of data augmentation methods, and their results on the test set.

| Training settings | Test Evaluation | | | |
| --- | --- | --- | --- | --- |
| | P | R | F1 | diff |
| CoNLL-2012 [b] | 64.4 | 51.8 | 57.4 | - |
| + ONP + RSM | 65.3 | 53.7 | 58.9 | +1.5 |
| + ONP + MCM | 65.0 | 53.2 | 58.5 | +1.1 |
| + ONP + BT | 64.9 | 53.5 | 58.6 | +1.1 |
| + ONP + CSA | 64.6 | 52.7 | 58.0 | +0.6 |
| + RSM + MCM | 65.1 | 53.6 | 58.7 | +1.4 |
| + RSM + BT | **65.4** | **53.8** | 59.0 | **+1.6** |
| + MCM + BT | 64.9 | 52.9 | 58.2 | +0.8 |
| + MCM + CSA | 65.1 | 53.4 | 58.6 | +1.2 |
| + BT + CSA | 64.6 | 52.4 | 57.8 | +0.4 |

Table 7: AZP resolution training settings with different combinations of data augmentation methods, and their results on the test set.

# 6 Conclusion

We tested five methods to detect and generate AZPs. We suggested a simple and effective way to resolve AZPs in the automatically generated samples. We evaluated the augmented data on the AZP identification and resolution tasks for Arabic. Training the models with the augmented data achieved new state-of-the-art results for both AZP tasks.

# Acknowledgements

We would like to thank the anonymous reviewers for their insightful comments and suggestions which helped to improve an early version of the paper.